\documentclass[pmlr]{jmlr}

\RequirePackage{graphicx}
 \usepackage{booktabs}
 \usepackage{multirow}
 \usepackage{xcolor}
\usepackage{longtable}
\makeatletter
\def\set@curr@file#1{\def\@curr@file{#1}} 
\makeatother
\usepackage{siunitx} 

\theorembodyfont{\upshape}
\theoremheaderfont{\scshape}
\theorempostheader{:}
\theoremsep{\newline}

\jmlrvolume{[VOLUME \# TBD]}
\jmlryear{2024}
\jmlrworkshop{Machine Learning for Healthcare}

\title[Fairness-Aware Multimodal Embedding]{Equitable Electronic Health Record Prediction with FAME: Fairness-Aware Multimodal Embedding}

\author{%
  \Name{Nikkie Hooman}\textsuperscript{1}\Email{nikkieh@smu.edu} \\
  \Name{Zhongjie Wu}\textsuperscript{1}\Email{zhongjiew@smu.edu} \\
  \Name{Eric C. Larson}\textsuperscript{1}\Email{eclarson@lyle.smu.edu} \\
  \Name{Mehak Gupta}\textsuperscript{1}\Email{mehakg@smu.edu} \\[4pt]
  \addr\textsuperscript{1}Department of Computer Science, Southern Methodist University, Dallas, TX, USA
}

\begin{document}

\maketitle

\begin{abstract}

Electronic Health Record (EHR) data encompasses diverse modalities—text, images, and medical codes—that are vital for clinical decision-making. To process these complex data, multimodal AI (MAI) has emerged as a powerful approach for fusing such information. However, most existing MAI models optimize for better prediction performance, potentially reinforcing biases across patient subgroups. Although bias reduction techniques for multimodal models have been proposed, the individual strengths of each modality and their interplay in both reducing bias and optimizing performance remain underexplored. In this work, we introduce FAME (Fairness-Aware Multimodal Embeddings), a framework that explicitly weights each modality according to its fairness contribution. FAME optimizes both performance and fairness by incorporating a combined loss function. We leverage the Error Distribution Disparity Index (EDDI) to measure fairness across subgroups and propose a sign-agnostic aggregation method to balance fairness across subgroups, ensuring equitable model outcomes. We evaluate FAME with BEHRT and BioClinicalBERT, combining structured and unstructured EHR data, and demonstrate its effectiveness in terms of performance and fairness compared to other baselines across multiple EHR prediction tasks.

\end{abstract}

\section{Introduction}






\label{sec:intro}
Healthcare decisions involve one or more physicians and health specialists that must interpret a myriad of data sources. Electronic health records (EHRs) offer a storage medium for this myriad of patient specific information across historical visits.  
Therefore, leveraging artificial intelligence (AI) in conjunction with EHRs holds transformative potential to improve healthcare. EHRs usually contains both structured (e.g., numerical, categorical) and unstructured data (e.g., text or image). Despite the routine use of EHR data to contextualize clinical history and inform medical decision-making, the majority of deep learning architectures are unimodal \citep{kline2022multimodal}---they only learn features from either structured or unstructured EHR data \citep{zhou2111radfusion}. Multimodal AI (MAI) has emerged as a technique for analyses that can combine multiple types of data (e.g., text, images, medical codes) simultaneously, rather than relying solely on one modality, to generate outputs or predictions. 


While multimodal AI (MAI) has gained traction in healthcare applications \citep{shaik2023survey, cui2024multimodal}, relatively little attention has been directed toward leveraging these advancements to promote fairness. Although existing work increasingly evaluates MAI models for fairness, few studies explicitly explore how the integration and interaction of multiple modalities can be harnessed to achieve more equitable outcomes. Our work addresses this research gap by proposing fairness-aware fusion techniques. Our proposed techniques weigh each modality based on its prediction and fairness performance to build efficient and fairer multimodal AI models. We assess fairness across attributes such as ethnicity, age, and insurance type using EDDI (Error Distribution Disparity Index) \citep{wang2024fairehr} across all subgroups. We propose a sign-agnostic aggregation method to combine EDDI across all subgroups that can be employed not only in the loss function for our model, but also as part of a feedforward weighting scheme to ensure modalities that promote fairness are given priority in the model. To test our approach, we utilize two language encoders designed for healthcare data: BEHRT and BioClinicalBERT. BEHRT is a transformer-based model that has performed well in analyzing structured longitudinal EHR data. BioClinicalBERT, on the other hand, is a specialized language model for unstructured clinical text. Combining these modalities ensures our model can process the myriad of input data from EHRs, while also providing a complex test case for our proposed EDDI weighting scheme.  




\textbf{Contributions:} 
\begin{enumerate}
    \item We propose fairness-aware multimodal embeddings (FAME), a method for fusing multiple modalities in EHR data using weighted aggregation.
    \item We introduce a method to derive fairness-aware weights using a sign-agnostic aggregation across EDDI values within a specified subgroup. Additionally, we implement a loss function that incorporates the aggregated EDDI to optimize the model while ensuring equality across subgroups.
    \item We demonstrate our method’s effectiveness through a series of experiments on three EHR prediction tasks, comparing it to other baselines and within an ablation of the model elements.
\end{enumerate}



\subsection*{Generalizable Insights about Machine Learning in the Context of Healthcare}
Existing multimodal AI models in healthcare primarily integrate multiple data types to enhance predictive performance, but they often do not explicitly account for how individual modalities contribute to fairness. While multimodal approaches can capture richer patient information, they typically treat different data sources as complementary inputs rather than considering their distinct roles in mitigating biases. Our work explores an alternative perspective by incorporating modality-specific weighting to promote fairness, demonstrating that structured and unstructured data can be leveraged not only for improving accuracy but also for reducing disparities across patient subgroups. This suggests that a more intentional approach to multimodal fusion—one that balances both predictive performance and fairness—may be beneficial in the development of equitable healthcare AI systems.

\section{Related Work}

In this section, we review existing bias mitigation methods, fusion methods, and fairness evaluation metrics in multimodal EHR models.

\subsection{Bias Mitigation in Multimodal EHR models}
The existence of biases and disparity across ethnicity, age, and other factors is a major concern in healthcare with the potential to be exacerbated by digital technologies and AI \citep{agarwal2023addressing}. Recent policy discussions have noted that with the growing impact of AI on business and society, it is even more critical to ensure that AI is fair, equitable, and treats all ``populations'' in an equivalent manner \citep{schwartz2022towards,sharman2022data,mehrabi2021survey}. This is especially important for healthcare, as health inequity can have life-altering consequences. 

Recent studies have proposed various strategies to reduce the impact of patient demographics and socioeconomic factors on AI model outputs. Adversarial learning frameworks, for example, have been employed to mitigate bias in clinical prediction by learning representations that are invariant to sensitive attributes \citep{liu2022mitigating, pfohl2021empirical, sivarajkumar2023fair}.  Similarly, contrastive learning techniques reduce contrastive loss between demographically based counterfactuals to promote performance parity across subgroups \citep{agarwal2024debias, wang2024fairehr}. 

In parallel, several multimodal EHR models have begun to incorporate interpretability mechanisms to better understand the contributions and interactions of individual modalities. For instance, \citet{lyu2023multimodal} proposed a multimodal transformer that jointly models clinical notes and structured data, emphasizing the distinct predictive strengths of each modality. \citet{tsai2020multimodal} introduced multimodal routing techniques to enhance both local and global interpretability, allowing examination of modality interactions during decision-making.

Despite these advances, most approaches apply debiasing techniques only after modality fusion, overlooking the unique fairness contributions of each modality. In contrast, our method explicitly evaluates and leverages the fairness of embeddings from individual modalities prior to fusion, enabling a more targeted and interpretable approach to multimodal bias mitigation.

\subsection{Multimodal fusion methods}

%

Several different strategies can be leveraged to fuse features from different modalities, including early fusion, late fusion, and joint fusion \citep{huang2020fusion, huang2020multimodal, zhou2111radfusion, baltruvsaitis2018multimodal}. Early fusion combines features from separate modalities at the input level. Late fusion trains separate models for each modality and aggregate the predicted probability from all single-modality models as the final prediction. In joint fusion, intermediate representation (embeddings) from unimodal models \citep{vaswani2017attention, devlin2019bert, dosovitskiy2020image, huang2019clinical, raffel2020exploring} are combined either through concatenation, addition, or MLP (multi-layer perceptron) fusion to obtain multimodal embeddings for each sample in the dataset. 


Recent advances in deep learning technologies have led to the development of complex multimodal AI models in healthcare \citep{kline2022multimodal, wang2024fairehr, shaik2023survey, cui2024multimodal,rahman2020integrating, radford2021learning}. Most of these existing techniques use multimodal fusion techniques like joint fusion or late fusion to combine embeddings and train models on performance-related metrics such as binary-cross entropy loss to produce high-performance models.

In our work, we focus on joint fusion and modify it to implement fairness-aware joint fusion. In existing joint fusion methods, embeddings from all modalities are either concatenated or averaged without weighting, which limits the utilization of the unique strengths of each modality. In our proposed method, FAME, we use fairness-metric information to weigh the unimodal embeddings before fusing them. 
This can help to control the contribution of each modality based on its observed fairness and leverage the joint fusion mechanisms not only to facilitate strong prediction but also to facilitate fair outcomes. 

\subsection{Fairness evaluation in Multimodal EHR models}
Traditional fairness metrics like equalized odds and disparity index assess fairness for different subgroups in each demographic category \citep{hardt2016equality}. According to these metrics, a model is considered fair if the error rate, like the true positive rate and false positive rate, across all subgroups is consistent. This measure ignores the relative size of subgroups and imbalanced outcomes in health data. From the wide range of metrics used in the community \citep{castelnovo2022clarification}, we use EDDI, specifically designed for datasets where subgroups and outcomes are imbalanced. Error Distribution Disparity Index (EDDI) \citep{hardt2016equality} measures the difference in error rates (proportion of incorrect prediction) between the privileged and unprivileged groups for a given prediction task. EDDI can be positive or negative depending on if subgroup error rate is higher or lower than the overall error rate, respectively. To obtain one EDDI value per category (e.g., race, insurance type) that ensures equality across all subgroups, we propose a sign-agnostic aggregation method to aggregate EDDI across each subgroup (e.g., White, Asian, Black) in a demographic category (e.g., race).

\section{Methods}

\subsection{Problem Formulation}
We define a multimodal dataset $D={(X_0,X_1,...X_m...,X_M)}_{m=1}^{M}$, where $M$ is the number of modalities and $X_m$ is feature data for all data points in the $m^{th}$ modality. Our objective is to develop an effective and fair multimodal EHR model $f_{multi}: {(X_0,X_1,...X_m...,X_M)}_{m=1}^{M} \rightarrow Y$, where $Y\in (0,1)$ are the output binary labels. While training $f_{multi}$, we aim to effectively combine multiple modalities to improve prediction performance while also ensuring equality among all subgroups, $S$.  Though our proposed methods can be applied to any number of modalites. In this work, we use $M=3$ (i.e., three modalities) where $X_d$ is demographic data, $X_s$ is longitudinal clinical data (structured), $X_n$ is longitudinal clinical notes (unstructured). We use a subset of demographic data as a set of sensitive attributes $S \subset X_d$. The subgroups, $S$, are selected using attributes identified as sensitive and, therefore, requiring fair outcomes across each group. The selection of these sensitive groupings is discussed in more detail in Section \ref{sec::sensitive_group}. 


Our proposed approach uses the joint fusion technique to combine multiple modalities. Joint (or intermediate) fusion trains a decision-making model using extracted features from single-modality models while propagating the loss back to the unimodal feature-extracting models \citep{zhou2111radfusion}. These extracted features are intermediate $n$-dimensional latent embeddings, $z_m$, that can be learned using any neural network. The input to each single-modality (or unimodal) model is only one type of data, $X_m$. These latent embeddings, $z_m$ are then fused to predict the outcome.


We define this process more mathematically below, where each latent embedding is denoted as $z$ and operations within the network are denoted by the function $f$. We define each model as:

$$
z_m = f_m(X_m), \qquad z_{fuse}=f_{fuse}(z_1, \dots, z_m), \qquad y=f_{cls}(z_{fuse})
$$

\noindent where $f_m$ is the $m^{th}$ unimodal model, $f_{fuse}$ is the fusion (aggregation) method used to fuse latent embeddings from all modalities, and $f_{cls}$ maps the fused information to a binary class variable, $y$. Combining all the operations above, we can define our multimodal model $f_{multi}$ as:

$$
y=f_{multi}(X_1, \dots, X_m) = f_{cls}\left( f_{fuse}\left(  f_1(X_1),  \dots, f_m(X_m)  \right) \right)
$$

\noindent One simple example of a fusion method, $f_{fuse}$, could be taking the average of each modality:

\begin{equation}\label{eq:1}
z_{fuse} \rightarrow \frac{1}{M} \sum_{m=1}^{M} z_m
\end{equation}


 

\begin{figure}[t]
  \centering
  \includegraphics[width=\linewidth,keepaspectratio]{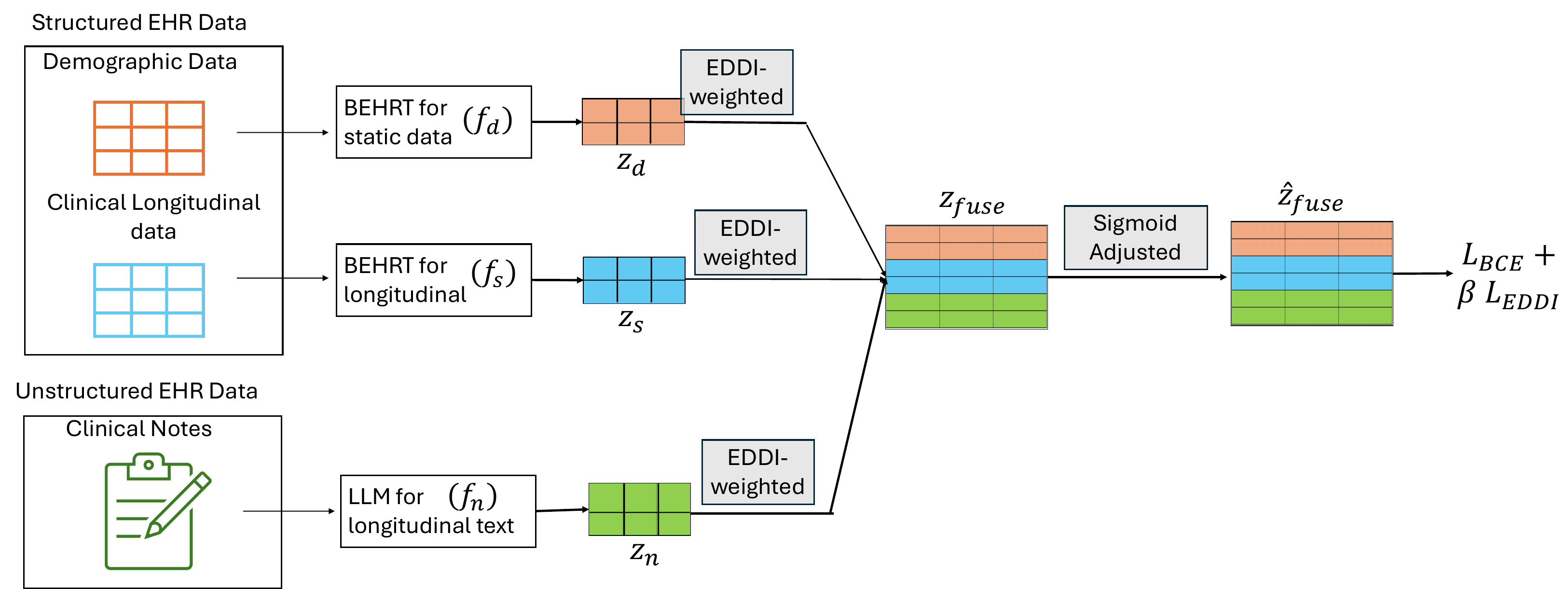}
  \caption{Model}
  \label{fig:model}
\end{figure}

\subsection{Fairness-aware Multimodal Fusion}
In the joint fusion discussed in Equation \ref{eq:1}, each modality is weighted equally towards fused embeddings $z_{fuse}$. We propose to weigh embeddings from each modality before combining them for multimodal fusion. Our goal is to find the relative weights of each modality towards the multimodal fusion based on its fairness performance. 

\begin{equation}\label{eq:wemb}
    z_{fuse}^{(t)} = \sum^{M}_{m=1}\hat{w}_{m}^{(t)} \cdot (z_{m}\cdot W_{m})
\end{equation}
where $z_{fuse}^{(t)}$ is the multimodality embedding at iteration $t$ of training, and $W_{m}$ is a trainable weight matrix that projects the $m^{th}$ modality. Finally, $\hat{w}_m^{(t)}$ is a gating vector in $t^{th}$ iteration for the $m^{th}$ modality that is influenced by fairness. This vector is defined more specifically below. Intuitively, this vector serves to scale dimensions within modalities that promote fairness and is recomputed at each iteration, $t$. 

\textbf{EDDI-weighted Fusion:}
To measure fairness for each modality, we compute the EDDI value across training iterations. EDDI can be computed by taking the mean across all subgroups, as proposed by \cite{wang2024fairehr}:

\begin{equation}
\label{eq:eddi_orig}
    EDDI = \frac{1}{|S|} \sum_{s\in S} \frac{ER_s - OER}{\max(OER,1-OER)}
\end{equation}

\noindent where $ER_s$ defines the error rate for subgroup $s$ (eg. White, Black, Asian) belonging to a sensitive attribute (eg. race) and $OER$ is the overall error rate for the dataset (i.e., the expected error rate). Error rate is defined in the customary manner as the percentage of predicted class labels that do not match the ground truth label. A potential downside of equation \ref{eq:eddi_orig} is that it adds all positive and negative EDDI values together, which can misrepresent the overall EDDI value for a sensitive attribute. To mitigate this, we propose to combine EDDI-values for all subgroups (eg. White, Black, Asian. etc) in a sensitive attribute (eg. race) using a sign-agnostic method. In this method, we take mean of the root of the sum of squares values of individual $EDDI_s$ for each subgroup. This helps ensure equitable fairness estimation across all demographic subgroups irrespective of their EDDI being positive or negative.

\begin{equation}
    EDDI_s =  \frac{ER_s - OER}{\max(OER,1-OER)}
\end{equation}
\begin{equation}\label{eq:rmse}
    EDDI_{S} = \frac{1}{|S|} \sqrt{\sum_{s\in S} EDDI_s^2}
\end{equation}

We then take the mean of the $EDDI_{S}$ value across all sensitive attributes (like, race, age, insurance) to obtain overall EDDI across each modality. Based on this EDDI, we calculate the weight parameter $w_m$ for each modality that modulates each embedding vector dimension through element-wise scaling (i.e., an element-wise scaling function). Embeddings from multiple modalities are adjusted using the following: 

\begin{equation}\label{eq:eddi}
    w_m^{(t)} = w_m^{(t-1)} + \gamma\cdot \left(\underset{m \in M}{\max}(EDDI_m^{(t)})-EDDI_m^{(t)}\right)
\end{equation}
where $w_m^{(t)}$ is the weight in $t^{th}$ iteration, $\gamma$ is the learning rate, and $EDDI_m^t$ is the mean EDDI value across all sensitive attributes for $m$ modality. A lower value of EDDI reflects a group with more fair treatment of groups. $\gamma$ controls how much $w_m^{(t)}$ should change each iteration based on the EDDI value for that iteration. The initial weight for each modality is assigned equally, where $w_m^{(0)}=\frac{1}{M}$. Therefore, when $\gamma$ is $0$ the result is that equal weight is given to each modality for all iterations, which reduces to average joint fusion given in Equation \ref{eq:1}. By subtracting $EDDI_m$ from $\max(EDDI_m)$ we ensure that modality with lower $EDDI_m$ receives more weight. Lastly, we take the normalized weight value across all modalities such that the sum of weights for all modalities is always unity. This $\hat{w}_m^{(t)}$ value is used in Equation \ref{eq:wemb} to obtain the weight modulated modalities.

\begin{equation}\label{eq:eddi_w}
    \hat{w}_m^{(t)}=\frac{w_m^{(t)}}{\underset{m \in M}{\sum}w_m^{(t)}} 
\end{equation}

For every modality we attach a light-weight classification head—a single linear layer followed by a sigmoid—to its embedding $z_m$.  
These heads are \emph{not optimised}: their outputs are excluded from the loss, so they receive no gradient updates and remain effectively fixed.  
At the end of each epoch we run them (in \texttt{torch.no\_grad()} mode) on the training data to obtain predictions $\hat{y}_m$, from which subgroup error rates $\mathrm{ER}_s$ and the modality-level $\mathrm{EDDI}^{(t)}_m$ are computed.  
Thus they function solely as fairness probes supplying the $\mathrm{EDDI}^{(t)}_m$ values used in Eq.~\eqref{eq:eddi}.

\textbf{Sigmoid-weighted Feature Selection:}
As an extension to the proposed EDDI-weighted fusion, we also propose a feature gating mechanism to help promote more fair individual features within each modality. After obtaining the EDDI-weighted embedding for each modality $z_{mw} = \hat{w}_{m} \cdot (z_{m}\cdot W_{m})$, we use a sigmoid activation to adjust embeddings by learning a weight parameter $\sigma(\cdot)$ that modulates each dimension of the embedding vector through element-wise scaling. For a given iteration, $t$, we first concatenate each $z_{mw}^{(t)}$ for all $m \in M$ into the vector $z_{concat}$ and take the dot product of sigmoid layer weights with each unimodal embedding:

\begin{equation}\label{eq:concat}
    z_{concat}^{(t)} = \text{concat}( z_{1w}^{(t)},\dots ,  z_{mw}^{(t)})
\end{equation}
\begin{equation}\label{eq:sig}
    \hat{z^t} = \sigma(W) \odot z^t_{\text{concat}}
\end{equation}

\noindent where $\hat{z}^t$ is the sigmoid adjusted embedding to modulate features from each modality and $W$ is the trainable weight vector,and $z_{concat}$ is the 768-D vector obtained after concatenating 256-D vector from each modality. To incorporate into our final model, we simply replace $z_{fuse}$ from Eq. \ref{eq:wemb} with $\hat{z^t}$ in the classification layer.  We later use the sigmoid output to analyze which features from each modality are weighted more or less for the final prediction by aggregating weights for each 256 features in 768-D $W$ vector. 

\subsection{Loss Functions}
Joint fusion allows the gradient update to flow through all unimodal and fused models together to train the whole model on the desired outcome. We use a combination of binary-cross entropy and EDDI loss to optimize our fusion model $f_{total}$
\begin{equation}\label{eq:loss}
    \mathcal{L}_{total} = \mathcal{L}_{BCE} + \lambda\cdot \mathcal{L}_{EDDI}
\end{equation}
where $\mathcal{L}_{BCE}$ is binary cross-entropy loss, $\mathcal{L}_{EDDI}$ is the mean of $EDDI_S$ (Equation \ref{eq:rmse}) across all sensitive attributes, and $\lambda$ is the hyperparameter to control the contribution of each loss.

\subsection{Model for Multimodal Fusion}
We will use different transformer-based models tailored for structured and unstructured data in our multimodal architecture.

 \textbf{BEHRT for Demographic and Longitudinal Structured Data:} BEHRT \citep{li2020behrt} is a transformer-based model specifically designed to process structured electronic health record (EHR) data, capturing the temporal, demographic, and clinical context of patient histories. It utilizes embedding layers for categorical values such as disease codes, age categories, and demographic data. These embeddings are combined with positional and segment embeddings to preserve temporal information about visits. We will utilize BEHRT model in our code for structured clinical and demographic data. We will extend the BEHRT model to include numerical values from patients' labs and vitals. For our joint fusion method, we will use $[CLS]$ embedding from last layer as the representative embedding for each patient's structured clinical and demographic data.



\begin{table}[t]
\centering
\caption{Outcome Label Distribution}
\label{tab:outcome_distribution}

\begin{tabular}{@{}lccc@{}}
\toprule
Outcome                 & Total Patients & Positive Cases & Percentage (\%) \\ \midrule
Short-Term Mortality    & 33,721         & 3,417          & 10.13            \\
Length of Stay(LOS)   & 33,721         & 4,991          & 14.80            \\
Mechanical Ventilation    & 33,721         & 30,356         & 90.02            \\ \bottomrule
\end{tabular}
\end{table}

\textbf{BioClinicalBERT for Unstructured Text Data:}
We will use BioClinicalBERT \citep{alsentzer2019publicly} to process unstructured clinical data from electronic health records (EHR). We began by preprocessing the dataset, where the text is divided in to 512 tokens each, followed by embedding extraction via a fine-tuned BioClinicalBERT model. [CLS] embedding from the last layer is combined using an MLP to obtain one patient-level embedding for use in the joint-fusion framework.



 \textbf{Multimodal Fusion:} We will combine embeddings from three data modalities — demographic ($z_d$), longitudinal structured clinical ($z_s$), and unstructured clinical notes ($z_n$) — to implement our fairness-aware joint fusion framework. 
 We will bring embeddings from all three unimodals into same $n-dimensional$ before applying fairness-aware weighting.

\section{Cohort}
\subsection{Cohort Selection}
Our cohort selection steps followed the benchmark by \cite{wang2020mimic} where we used only first ICU visit for all patients above 15 years of age. In all cases, we use the first 24 hours
of a patient’s data, only considering patients with at least 30 hours of present data. This 6 hour gap time is critical to prevent temporal
label leakage. For structured data we followed Feature Set C from \cite{purushotham2018benchmarking} to select 136 features from MIMIC-III which is a superset of features selected by \cite{wang2020mimic}. We aggregated temporal features into 2-hour bins.  Additionally, textual embeddings from BioClinicalBERT were incorporated to capture information from physician notes, nursing progress records, and radiology reports. 

\subsection{Prediction Tasks}
We use our data to predict three binary classification tasks: In-ICU Mortality, Length of Stay (LOS) $\ge$ 7, and Mechanical Ventilation.

The In-ICU Mortality label was extracted from the DEATHTIME field in the MIMIC-III dataset, where a non-null entry signifies a death event. The length of stay (LOS) prediction task aims to classify whether a patient’s ICU stay will exceed three days using the first 24 hours of clinical data. The label was derived from the INTIME and OUTTIME timestamps in the ICUSTAYS table of the MIMIC-III dataset. Mechanical ventilation is a critical intervention in ICU patients, often indicating severe respiratory distress or failure. We identified ventilation-related item IDs from the CHARTEVENTS and PROCEDUREEVENTS\_MV tables, including ventilator settings, oxygen therapy usage, and extubation events. Additionally, procedural records indicating intubation or tracheostomy were incorporated to refine label accuracy.

\subsection{Sensitive Attributes}
\label{sec::sensitive_group}
We will use sensitive attributes - ethnicity (in the MIMIC dataset, both race and ethnicity are recorded under ethnicity), insurance, and age. Each attribute is further divided into subgroups with ethnicity having 5 subgroups - White, Black, Hispanic, Asian and Other, insurance which can act as a proxy for socio-economic status has 5 subgroups - Medicare, Private, Medicaid, Goverment, and Self-pay, and age divided into subgroups - 15-29, 30-49, 50-69, and 70+. We have shared Table \ref{tab:feature_distribution}, which shows the distribution of all subgroups in the sensitive attributes. 
\begin{table}[t]
\small
\centering
\caption{Sensitive Attribute Distribution}
\label{tab:feature_distribution}

\begin{tabular}{@{}lccc@{}}
\toprule
Feature      & Subcategory & Count  & Percentage (\%) \\ \midrule
Age Bucket   & 15--29      & 1,832  & 5.4             \\
             & 30--49      & 5,729  & 17.0            \\
             & 50--69      & 13,344 & 39.6            \\
             & 70+         & 12,816 & 38.0            \\ \midrule
Ethnicity    & White       & 23,887 & 70.8            \\
             & Black       & 2,567  & 7.6             \\
             & Hispanic    & 1,076  & 3.2             \\
             & Asian       & 670    & 2.0             \\
             & Other       & 5,521  & 16.4            \\ \midrule
Insurance    & Medicare    & 17,163 & 50.9            \\
             & Private     & 12,151 & 36.0            \\
             & Medicaid    & 2,889  & 8.6             \\
             & Government  & 1,060  & 3.1             \\
             & Self Pay    & 458    & 1.4             \\ \bottomrule
\end{tabular}

\end{table}

\section{Experiments and Results}

\subsection{Implementation Details} 

To implement multimodal models we project the [CLS] embeddings from all three modalities into a shared 256-dimensional space. In the initial iteration, each modality is assigned equal weight, and their embeddings are concatenated to form a 768-dimensional representation. In subsequent iterations, we use the mean EDDI weights computed across three sensitive attributes—ethnicity, insurance status, and age—to weight each modality. These EDDI weights are updated at each iteration using a learning rate ($\gamma$) of 0.5. The final concatenated embedding is passed through a linear classification layer with a hidden dimension of 512 and a dropout rate of 0.2 to predict three binary outcomes.

To address class imbalance in the prediction tasks, we use a weighted binary cross-entropy loss function, where class weights are determined via the Inverse of Number of Samples (INS) method. Specifically, we employ PyTorch’s BCEWithLogitsLoss, assigning higher weights to minority classes to mitigate skewed distributions.

The model is optimized using the AdamW optimizer with a learning rate of $1 \times 10^{-5}$ and a weight decay of 0.01 for regularization. We also incorporate the ReduceLROnPlateau scheduler from to adaptively reduce the learning rate based on validation performance.

Training is performed on an 80-20 split of the data, with 80\% used for training and 20\% for testing. Additionally, 5\% of the training data is held out for validation. Early stopping is applied with a patience of 5 epochs, terminating training if validation loss does not improve across 5 consecutive epochs. The model checkpoint with the lowest validation loss is selected for final evaluation on the test set.
All hyper-parameters were tuned via grid search on the validation split; see Appendix~\ref{app:hparams} for the full search ranges and chosen values.

\subsection{Baseline Comparison}
We compare our proposed model against several state-of-the-art methods designed to mitigate bias in EHR prediction tasks. Evaluations are conducted on both predictive performance and fairness metrics to provide a comprehensive assessment of model effectiveness.

Demographic-free Classification
(DfC): It is an established method to reduce the impact of sensitive attributes on outcome is by unawareness. DfC is based on the concept of unawareness, which states that if demographic features are not included in input data, it should have minimal effect on output.

AdvDebias \citep{zhang2018mitigating,yang2023adversarial}: It uses adversarial training to debias patient representations. It trains an adversary model such that the model cannot correctly classify patient's sensitive attributes from the patient representations, thus debiasing patient representations. A classifier is trained on debiased patient representation to improve fairness. 

Fair Patient Model (FPM) \citep{sivarajkumar2023fair}: FPM employs a Stacked Denoising Autoencoder and a weighted reconstruction loss for equitable patient
representations. 

FairEHR-CLP \citep{wang2024fairehr}: FairEHR-CLP uses contrastive learning between patient data and synthetically generated counterfactual samples with different sensitive attributes but the same medical histories as the original sample. By reducing contrastive loss between original and counterfactuals, the aim is to reduce bias in the multimodal EHR model.

\begin{table*}[t]
\small
    \centering
    \caption{Baseline comparison using performance and fairness evaluation across three prediction tasks. We report average results over 5 runs. EDDI, and EO are averaged over three sensitive attributes. Bold indicates best results.}
    \label{tab:results}
    \resizebox{\linewidth}{!}{%
    \begin{tabular}{llcccccc}
        \toprule
        Task & Model & AUROC ($\uparrow$)  & AUPRC ($\uparrow$) & EDDI($\%$) ($\downarrow$)& EO($\%$)($\downarrow$)\\
        \midrule
        \multirow{5}{*}{In-ICU Mortality} 
        & DfC                        & 0.90 & 0.71 & 0.79 & 5.80 \\
        & AdvBias                    & 0.93 & 0.75 &  2.40 & 9.91 \\
        & FPM                        & 0.93 & 0.74 & 6.94 & 14.10 \\
        & FairEHR-CLP                & 0.92 & 0.74 &  8.84 & 16.10 \\
          
        & FAME (Ours)  & \textbf{0.94} & \textbf{0.82} & \textbf{0.44} & \textbf{4.25} \\    
        \midrule
        \multirow{5}{*}{LOS $\ge$ 7} 
        & DfC                        & 0.98 & 0.91 & 0.55 & 2.83 \\
        & AdvBias                    & 0.96 & 0.85 &  2.39 & 5.70 \\
        & FPM                        & 0.96 & 0.86 & 3.43 & 11.10 \\ 
        & FairEHR-CLP                & 0.96 & 0.85 &  3.85 & 12.75 \\
       
        & FAME (ours) & \textbf{1.00} & \textbf{1.00} & \textbf{0.02} & \textbf{0.06} \\       
        \midrule
        \multirow{5}{*}{Mechanical} 
        & DfC                        & 0.78 & 0.97 & \textbf{1.29} & 2.58 \\
       \multirow{5}{*}{Ventilation}  
        & AdvBias                    & 0.84 & 0.97 & 1.73 & 15.70 \\ 
        & FPM                        & 0.83 & 0.97 & 7.51 & 13.50 \\  
        & FairEHR-CLP                & 0.84 & 0.97 & 9.67 & 16.40 \\  
       
        & FAME (Ours)  & \textbf{0.84} & \textbf{0.97} & 2.77 & \textbf{0.55} \\      
        \bottomrule
    \end{tabular}%
    }
\end{table*}

\begin{table*}[t]
\small
    \centering
    \caption{Ablation analysis across modalities and model components using performance and fairness evaluation across three prediction tasks. We report average results over 5 runs. EDDI, and EO are averaged over three sensitive attributes. Bold indicates best results.}
    \label{tab:ablation}
    \resizebox{\linewidth}{!}{%
    \begin{tabular}{llcccccc}
        \toprule
        Task & Model & AUROC ($\uparrow$)  & AUPRC ($\uparrow$) & EDDI($\%$) ($\downarrow$)& EO($\%$)($\downarrow$)\\
        
       
        
        \midrule
        \multirow{6}{*}{In-ICU Mortality} 
        
        & BEHRT                      & 0.83 & 0.40 & 2.08 & 5.20 \\    
        & BioClinicalBERT            & 0.93 & 0.78 &  0.82 & 4.80 \\
        & Average Fusion  & 0.93 & 0.74 & 1.39 & 6.91 \\     
        & Sigmoid-only  & 0.86 & 0.40 & 6.82 & 15.53 \\    
        & EDDI-only     & 0.92 & 0.69 & 1.66 & 7.00 \\          
        & FAME (Ours)  & \textbf{0.94} & \textbf{0.82} & \textbf{0.44} & \textbf{4.25} \\      
        \midrule
        \multirow{6}{*}{LOS $\ge$ 7} 
       
        & BEHRT                      & 0.76 & 0.35 & 1.34 & 4.24 \\    
        & BioClinicalBERT            & 0.96 & 0.85 &  0.66 & 4.01 \\ 
        & Average Fusion  & 0.96 & 0.85 & 0.34 & 3.97 \\      
        & Sigmoid-only & 1.00 & 0.97 & 0.09 & 0.20 \\       
        & EDDI-only     & 1.00 & 0.99 & 0.07 & 0.06 \\    
        & FAME (ours) & \textbf{1.00} & \textbf{1.00} & \textbf{0.02} & \textbf{0.06} \\       
        \midrule
        \multirow{6}{*}{Mechanical} 
        
        & BEHRT                      & 0.80 & 0.97 & 4.72 & 8.23 \\    
        \multirow{6}{*}{Ventilation} 
        & BioClinicalBERT            & 0.78 & 0.97 &  \textbf{1.98} & 2.75 \\ 
        & Average Fusion  & 0.83 & 0.97 & 3.11 & 6.93 \\     
        & Sigmoid-only & 0.67 & 0.94 & 4.59 & 8.79 \\    
        & EDDI-only     & \textbf{0.88} & \textbf{0.98} & 4.96 & 7.17 \\    
        & FAME (Ours)  & 0.84 & 0.97 & 2.77 & \textbf{0.55} \\     
        \bottomrule
    \end{tabular}%
    }
\end{table*}

\subsection{Ablation Analysis}
As part of the ablation analysis, we will compare our model with below modifications:

BEHRT: We will use unimodal BEHRT only for structured clinical data modality $(X_s)$ to predict three prediction tasks.

BioClinicalBERT: We will use unimodal BioClinicalBERT only for unstructured clinical text modality $(X_n)$ to predict three prediction tasks.

Average Fusion: Multimodal model with three modalities combined using fusion method in Eq. \ref{eq:1}

Sigmoid-only: Our proposed multimodal model with three modalities combined using concatenated fusion method in Eq. \ref{eq:concat} and sigmoid function in Eq. \ref{eq:sig}.

EDDI-only: Our proposed multimodal model with three modalities combined using EDDI-weighting fusion in Eq. \ref{eq:wemb} with weights from Eq. \ref{eq:eddi} and Eq. \ref{eq:eddi_w} and no Sigmoid-weighted feature selection.


\subsection{Evaluation Metrics} 
We evaluate the model using both standard predictive performance metrics and fairness-aware assessments. We use Area Under the Receiver Operating Characteristic Curve (AUROC) and Area Under the Precision-Recall Curve (AUPRC) to assess overall model performance and precision-recall trade-offs. To assess fairness across different demographic groups, we employed EDDI and Equalized Opportunity (EO). To calculate EO we took mean of aggregated True positive rate (TPR) and False positive rate (FPR). To calculate aggregated TPR and FPR, we first calculate the absolute pairwise difference of TPR and FPR between each subgroup of a sensitive attribute and then take the mean over three attributes \citep{wang2024fairehr}. We also report EDDI by first aggregating at subgroup-level (Eq. \ref{eq:rmse}) and then taking the arithmetic mean over all sensitive attributes. All these fairness metrics need to be lower to demonstrate the fairness of a model.

\section{Results} 

\subsection{Baseline Comparison}
Table~\ref{tab:results} presents a comparative evaluation of our proposed model against several state-of-the-art baselines using four key metrics: AUROC, AUPRC, Error Distribution Disparity Index (EDDI), and Equalized Odds (EO). Although DfC consistently underperforms in AUROC and AUPRC, it achieves lower bias compared to other baselines. This indicates that completely omitting demographic features can help reduce bias, but often at the significant cost of predictive performance. These findings suggest that rather than exclusion, a more effective strategy may be to regulate the influence of demographic features to strike a balance between fairness and accuracy. Our proposed model excels in AUROC and AUPRC compared to all baselines and has reduced bias compared to baselines in most settings. It shows the importance of including all modalities to improve performance but also controlling their contribution to produce equitable outcomes.

\begin{table*}[t]
    \centering
    \caption{Fairness Results for all sensitive attributes compared across different components of our proposed model.}
    \label{tab:eddi_results}
    \resizebox{\linewidth}{!}{%
    \begin{tabular}{lccccccccc}
        \toprule
         Model &  EDDI ($\%$) ($\downarrow$)& EDDI($\%$)($\downarrow$) & EDDI ($\%$) ($\downarrow$)&EO ($\%$)($\downarrow$)& EO ($\%$)($\downarrow$) & EO ($\%$)($\downarrow$) \\
         Model &   (Age) &  (Ethnicity) &  (Insurance)& (Age)&  (Ethnicity) &  (Insurance) \\
       
         \midrule
         Sigmoid-Only               & 1.93 & 9.06 & 1.89  & 3.35 & 19.91 & 4.18 \\
         EDDI-only                  & \textbf{0.99} & 5.31 & \textbf{1.04}  & \textbf{1.34} & 10.89 & \textbf{1.99} \\
         FAME                       & 1.85 & \textbf{0.47} & 1.10 & 1.57 & \textbf{0.8} & 2.48 \\
        \bottomrule
    \end{tabular}%
    }
\end{table*}

\subsection{Ablation Analysis:} 
\textbf{Modality Analysis:} As part of our ablation study in Table \ref{tab:ablation}, we first evaluate the predictive effectiveness of individual modalities by training two unimodal models: BEHRT, which uses only structured clinical data, and BioClinicalBERT, which relies solely on clinical notes. BioClinicalBERT outperforms BEHRT by 12\% in AUROC and 77\% in AUPRC and also reduces bias by lower EDDI by 56\% and EO by 26\%, highlighting the richness and diversity of information embedded in clinical text. Clinical notes often capture not only clinical observations but also implicit demographic and social context, contributing to their predictive strength. When evaluated for fairness, our proposed model shows lower bias evident from lower EDDI and EO scores in most settings, over BioClinicalBERT (best performing unimodal), demonstrating the benefit of leveraging complementary information across modalities.

\textbf{Component Analysis:} Our ablation analysis of model components (Table~\ref{tab:ablation}) reveals that our proposed model FAME, which incorporates EDDI-based modality weighting, outperform the average fusion baseline that weighs all modalities equally. Specifically, FAME achieves overall improvements of 3\% in AUROC, and 9\% in AUPRC and significant bias reductions of 57\% and 76\% in EDDI and EO, highlighting the effectiveness of EDDI-weighted modality aggregation. Furthermore, the Sigmoid-only model exhibits a greater increase in bias compared to the EDDI-only model when both are evaluated against the full model, reinforcing the importance of EDDI-based weighting. The full model consistently yields the best fairness outcomes (lowest EDDI scores), suggesting that the combination of EDDI-weighted aggregation and Sigmoid-based feature selection provides complementary advantages for both performance and fairness.


\subsection{Fairness Comparison Across Sensitive Attributes:} In Table \ref{tab:eddi_results} we compare EDDI, and EO separately for our three sensitive attributes across different variations of our model. We average EDDI, and EO across three tasks. We observe that EDDI-only and FAME has consistently has lower EDDI and EO across all sensitive attributes compared to Sigmoid-only, reinforcing the importance of EDDI-weighting of individual modalities. Out of three sensitive attributes age and insurance (a proxy for socio-economic status) are more biased compared to ethnicity (also includes race).

\subsection{Sensitivity Analysis:} 
\begin{figure}[btp]
   \centering 
   \includegraphics[scale=0.3]{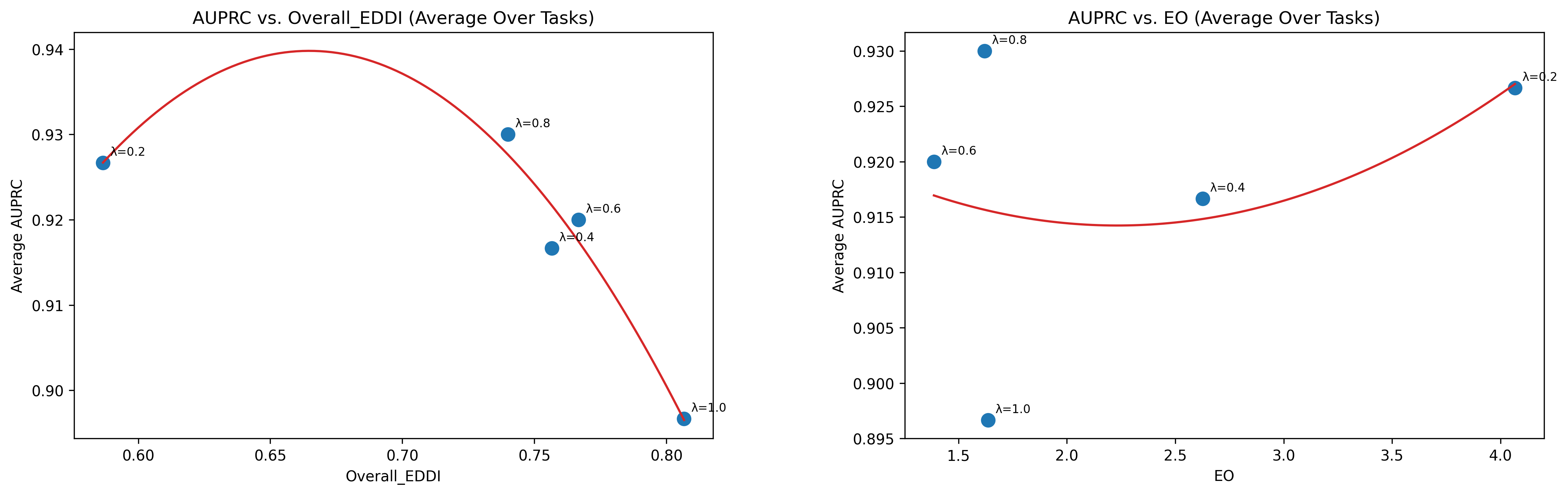} 
   \caption{Sensitivity analysis to study effect of $\lambda$ on performance and fairness. We compare AUPRC vs EDDI, AUPRC vs EO average over all tasks.}
   \label{fig:sensi} 
 \end{figure}

We also conducted a sensitivity analysis of the hyperparameter $\lambda$ (Eq.\ref{eq:loss}) to examine the trade-off between the binary cross-entropy loss and the EDDI loss in shaping model performance. Figure \ref{fig:sensi} illustrates a general trend where both EDDI and EO decrease as AUPRC increases, highlighting the inherent tension between fairness and accuracy and demonstrating the role of $\lambda$ in balancing the two objectives. However, this trend is not strictly monotonic across all values, which is common in deep learning models due to regularization effects. Typically, performance follows a U-shaped curve with respect to the regularization parameter $\lambda$, where an optimal value of 0.8 lies in the middle, rather than showing a strictly increasing or decreasing trend.

 \begin{figure}[t]
   \centering 
   \includegraphics[scale=0.51]{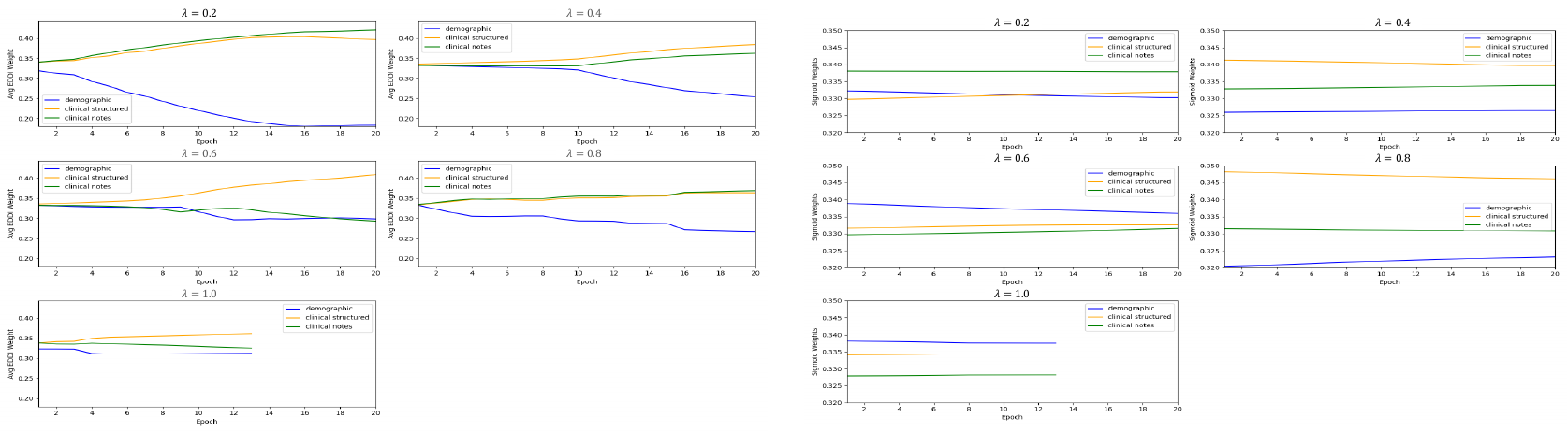} 
   \caption{Visualization of EDDI weights and Sigmoid weights changing over training iterations averaged across three tasks. We show results for different $\lambda$ values.}
   \label{fig:weights} 
 \end{figure} 
 
\subsection{EDDI-weights and Sigmoid-weights Visualization:}
Our proposed model is trained on joint loss of BCE and EDDI. To investigate how EDDI weight and Sigmoid feature selection evolve while our model is being optimized on joint loss we visualize changing weights over training epochs. In the left panel of Figure \ref{fig:weights} we illustrate how EDDI weights, computed using Eq.\ref{eq:eddi}, evolve during training. The right panel shows the progression of Sigmoid weights, derived from Eq.\ref{eq:sig}, over the course of model optimization. Both EDDI and Sigmoid weights are averaged across three tasks for visualization. Additionally, the Sigmoid layer weights are averaged over the 256-dimensional vectors corresponding to each modality, which are concatenated as described in Eq.~\ref{eq:concat}. 

The EDDI weights, initialized uniformly across modalities, gradually diverge during training, assigning higher weights to the structured clinical and unstructured text modalities, while the weight for the demographic modality consistently decreases across all settings. This trend is consistent with our baseline comparison, where removing the demographic modality from the DfC model improves EDDI performance. The Sigmoid feature selector continues to extract features from the demographic modality but assigns them lower importance compared to the other two modalities. EDDI weights for both structured and unstructured data increase over time, with the text modality receiving slightly higher weights in some settings ($\lambda$=0.2, 0.8). The Sigmoid layer, however, assigns greater emphasis to features from the structured modality (($\lambda$=0.4, 0.8)). At the optimal $\lambda$ value of 0.8 from our sensitivity analysis in Figure \ref{fig:sensi}, EDDI assigns nearly equal weights to the structured and text modalities, while the feature selection layer slightly favors the structured modality. Also, the anomaly behaviors at $\lambda$=1.0 for EDDI in Figure \ref{fig:sensi} can be attributed to higher demographic weight in the right panel of Figure \ref{fig:weights}.


\subsection{Interactions between
  \texorpdfstring{$\lambda$}{\(\lambda\)},
  EDDI weights and Sigmoid weights}

In the left panel of Figure \ref{fig:sensi}, we observe that $\lambda = 1.0$ yields the lowest AUPRC, as expected, and $\lambda = 0.2$ achieves higher AUPRC than $\lambda = 0.4$ and $\lambda = 0.6$, which aligns with our expectations. Interestingly, $\lambda = 0.8$ gives the highest AUPRC, though only marginally better than $\lambda = 0.2$. On the fairness side, while $\lambda = 1.0$ results in the worst fairness score, $\lambda = 0.2$ yields the best fairness outcome, which deviates from the expected trend. Nevertheless, $\lambda = 0.8$ performs better in terms of EDDI compared to $\lambda = 0.4$ and $\lambda = 0.6$, supporting the overall pattern. Looking at Figure \ref{fig:weights} for further insight, we see that $\lambda = 1.0$ assigns nearly equal EDDI weights across all three modalities, similar to the initial EDDI weights, suggesting that the model is underfitting and not effectively learning meaningful representations. The right panel of Figure \ref{fig:weights} also shows disproportionately high feature selection from the demographic modality, with weights again remaining close to the initial values assigned in the 0th iteration at $\lambda = 1.0$, further indicating poor learning behavior. This underfitting leads to suboptimal classification performance and contributes to the anomalous fairness outcomes observed.  

In the right panel of Figure \ref{fig:sensi}, AUPRC for $\lambda = 1.0$ is the lowest, as expected, with $\lambda = 0.2$ achieving higher AUPRC than $\lambda = 0.6$ and $\lambda = 0.4$, which aligns with expected behavior. The difference in AUPRC between $\lambda = 0.4$ and $\lambda = 0.6$ is minimal. On fairness metric $\lambda = 1.0$, $\lambda = 0.8$, and $\lambda = 0.6$ yield lower EO compared to $\lambda = 0.4$, and $\lambda = 0.2$. Notably, $\lambda = 0.8$ delivers the best overall performance in terms of both EO and AUPRC.  

Further supporting this, Figure \ref{fig:weights} shows that $\lambda = 0.8$ starts learning weights for each modality early in the training and results in higher EDDI weights for the structured and unstructured modalities compared to demographic, and it learns different weights for each modality giving higher weights to structured modality and least weight to demographic. Both observations are consistent with the strong performance observed at $\lambda = 0.8$. 
\section{Discussion}


The improved fairness and prediction performance results achieved by FAME highlight the limitations of relying on a single modality, even one as rich as clinical notes, and emphasize the value of integrating multiple modalities to capture complementary signals and mitigate bias. These findings reinforce the importance of explicitly accounting for modality-specific fairness contributions, as done through EDDI-based weighting. Compared to naive equal-weighted fusion, EDDI-weighted aggregation leads to more equitable outcomes across subgroups, as evidenced by reduced EDDI and EO scores.

Interestingly, while the model continues to access demographic information, minimizing its influence appears to support better fairness. The consistent reduction in demographic weights, observed through both EDDI and Sigmoid components, suggests that while such features may offer signal, their overemphasis can amplify bias. In contrast, structured and unstructured clinical data show increasing importance over training, with their divergent selection dynamics indicating complementary strengths of our model components.

The relatively higher weighting of the structured modality suggests a promising direction for future work: extracting both clinical and non-clinical information from unstructured text and integrating it into structured (tabular) formats. Such an approach could enhance not only model performance but also fairness by making valuable information from clinical notes more accessible to structured modeling techniques.





\paragraph{Limitations and Future Work}
While our study demonstrates the potential of fairness-aware multimodal fusion, there are a few limitations worth noting. First, we did not incorporate image modalities in our current experiments. Although the proposed framework is designed to support additional modalities, our evaluation was focused on structured, unstructured text, and demographic data. Second, our analysis considered a limited set of sensitive attributes—race, insurance status, and age—which may not fully represent the range of factors that contribute to disparities in healthcare. Lastly, the results are based on a single dataset, which may affect the generalizability of our findings across datasets.

As part of future work, we plan to extend our evaluation to datasets that include imaging data to better understand how visual information complements other modalities in enhancing both performance and fairness. We also aim to explore a broader range of sensitive attributes, including Social Determinants of Health (SDoH) identified through clinical notes, to capture a more complete picture of patient subgroups. Finally, we hope to validate our approach across multiple datasets to further examine how dataset characteristics influence model behavior and fairness outcomes.



\bibliography{sample}

\appendix

\section{Hyper-parameter Search Details}\label{app:hparams}

All key hyper-parameters were chosen via grid search on the validation split, jointly optimising predictive performance (binary cross-entropy) and fairness (EDDI). The ranges explored and the final choices are summarised in Table~\ref{tab:hparams}.  

\begin{itemize}
  \item \textbf{Learning rate} $\in\{1\times10^{-5},\,5\times10^{-6},\,2\times10^{-5}\}$ and
        \textbf{batch size} $\in\{8,\,16,\,32\}$ were tuned first to ensure stable transformer
        and fusion-layer training.  
        A learning rate of $1\times10^{-5}$ with batch size $16$ produced the best convergence
        and validation metrics.
  \item \textbf{Fairness weight} $\lambda\in\{0.2,0.4,0.6,0.8,1.0\}$ controls the EDDI loss
        term (Fig.~2, Sec.~6.4).  
        $\lambda=0.8$ gave the optimal AUPRC–vs.–fairness trade-off.
  \item \textbf{Weight-update rate} $\gamma\in\{0.5,1.0\}$ (Eq.~6) sets the step size for
        modality weights (clipped at $\pm0.05$ per epoch).  
        $\gamma=1.0$ yielded the most consistent fairness gains without oscillation.
  \item \textbf{L1 regularisation} of the sigmoid gates used
        $\alpha\in\{0.01,0.1\}$; $\alpha=0.01$ encouraged modest sparsity without hurting
        accuracy.
\end{itemize}

\begin{table}[h]
  \centering
  \begin{tabular}{@{}ll@{}}
    \toprule
    \textbf{Hyper-parameter} & \textbf{Search grid (bold = chosen)} \\
    \midrule
    Learning rate           & $5\times10^{-6},\;\mathbf{1\times10^{-5}},\;2\times10^{-5}$ \\
    Batch size              & $8,\;\mathbf{16},\;32$ \\
    $\lambda$ (EDDI loss)   & $0.2,\,0.4,\,0.6,\,\mathbf{0.8},\,1.0$ \\
    $\gamma$ (update rule)  & $0.5,\,\mathbf{1.0}$ \\
    L1 coefficient          & $\mathbf{0.01},\,0.1$ \\
    \bottomrule
  \end{tabular}
  \caption{Grid-search ranges and selected hyper-parameters.}
  \label{tab:hparams}
\end{table}
\section{Data Preprocessing}\label{sup:pre}

Preprocessing involved structured and unstructured data extracted from the MIMIC-III database. Each data type required tailored techniques to prepare it for model training while addressing the specific challenges inherent in clinical datasets.

\subsection{Structured Data Preprocessing}

For the structured data, we load core MIMIC-III datasets (\texttt{ADMISSIONS}, \texttt{PATIENTS}, and \texttt{ICUSTAYS}) and convert critical timestamp fields (e.g., \texttt{ADMITTIME}, \texttt{DISCHTIME}, \texttt{DEATHTIME}, \texttt{INTIME}, \texttt{OUTTIME}) to \texttt{datetime} format. Columns are renamed for consistency and ICU stays are merged with admissions and patient demographics to provide comprehensive clinical context, including outcomes such as short-term mortality (based on the presence of \texttt{DEATHTIME}) and hospital readmission within 30 days (computed by assessing the interval between consecutive ICU admissions). Age at ICU admission is calculated from date of birth and categorized into predefined buckets (15--29, 30--49, 50--69, 70--89), while ethnicity and insurance information are standardized using rule-based mappings. Categorical features (except gender) are one-hot encoded to facilitate model input. Finally, the structured dataset is filtered to retain only patients whose IDs are common with the unstructured data.

\subsection{Unstructured Data Preprocessing}
For the unstructured data, we process the NOTEEVENTS dataset by converting CHARTDATE to datetime and applying text cleaning procedures to remove extraneous characters, standardize abbreviations, and eliminate unnecessary whitespace. Concatenating notes aggregate clinical notes corresponding to each patient’s first ICU stay for a given hospital admission, and then split into chunks of approximately 512 tokens for compatibility with transformer-based models. The unstructured dataset is filtered to include only patients that overlap with the structured dataset. This integrated preprocessing pipeline ensures that structured (demographic, clinical, and laboratory) and unstructured (free-text clinical notes) data are aligned at the patient level, enabling a robust multimodal analysis.

\section{Mechanical ventilation Labels}\label{sup:vent}
Mechanical ventilation was identified by integrating data from the CHARTEVENTS and PROCEDUREEVENTS\_MV datasets. For CHARTEVENTS, we load the columns ICUSTAY\_ID, CHARTTIME, ITEMID, VALUE, and ERROR, filter out rows with null values or errors, and retain only those with predefined ventilation-related ITEMIDs (e.g., 720, 223848, 223849, 467). A helper function (determine\_flags) sets flags (mechvent, oxygentherapy, extubated, selfextubated) based on the recorded values. Similarly, the PROCEDUREEVENTS\_MV dataset is filtered for extubation-related ITEMIDs (after renaming STARTTIME to CHARTTIME) to mark extubation events, with extubated set to 1 and other flags set to 0. These signals are concatenated and merged with ICU stay data from ICUSTAYS (adding SUBJECT\_ID and HADM\_ID), and then aggregated by taking the maximum value across the flags for each ICU stay. The final binary label is computed as: \[mechanical\_ventilation = \max(mechvent, oxygentherapy, extubated, selfextubated)\] A patient is labeled as 1 if any of these flags is positive during their ICU stay.

\end{document}